\documentclass{article}
\usepackage{ijcai24}

\usepackage{times}
\usepackage{soul}
\usepackage{url}
\usepackage[hidelinks]{hyperref}
\usepackage[utf8]{inputenc}
\usepackage[small]{caption}
\usepackage{graphicx}
\usepackage{amsmath}
\usepackage{amsthm}
\usepackage{booktabs}
\usepackage{algorithm}
\usepackage{algorithmic}
\usepackage{multirow} 
\usepackage{bbm}
\usepackage{xspace}
\usepackage{arydshln} 
\usepackage[switch]{lineno}
\newcommand{\ie}{\textit{i.e.}}
\newcommand{\eg}{\textit{e.g.}}
\newcommand{\vs}{\textit{v.s.}}

\title{Probabilistic Contrastive Learning for Domain Adaptation}

\author{
 Junjie Li$^{1,2}$ \and Yixin Zhang$^{2}$ \and Zilei Wang$^{2}$\thanks{Corresponding Author} \and Saihui Hou$^{3}$ \and
 Keyu Tu$^{2}$ \and Man Zhang$^{1}$
 \affiliations
 $^{1}$Beijing University of Posts and Telecommunications\\
 $^{2}$University of Science and Technology of China\\
 $^{3}$Beijing Normal University\\
 \emails
   hnljj93@gmail.com,zhyx12@mail.ustc.edu.cn, zlwang@ustc.edu.cn, 
 housaihui@bnu.edu.cn,
 tky2017ustc\_dx@mail.ustc.edu.cn, zhangman@bupt.edu.cn 
}

\begin{document}

\maketitle


\begin{abstract}
Contrastive learning has shown impressive success in enhancing feature discriminability for various visual tasks in a self-supervised manner, but the standard contrastive paradigm (features+$\ell_{2}$ normalization) has limited benefits when applied in domain adaptation. We find that this is mainly because the class weights (weights of the final fully connected layer) are ignored in the domain adaptation optimization process, which makes it difficult for features to cluster around the corresponding class weights. To solve this problem, we propose the \emph{simple but powerful} Probabilistic Contrastive Learning (PCL), which moves beyond the standard paradigm by removing $\ell_{2}$ normalization and replacing the features with probabilities. PCL can guide the probability distribution towards a one-hot configuration, thus minimizing the discrepancy between features and class weights. We conduct extensive experiments to validate the effectiveness of PCL and observe consistent performance gains on five tasks, i.e., Unsupervised/Semi-Supervised Domain Adaptation (UDA/SSDA), Semi-Supervised Learning (SSL), UDA Detection and Semantic Segmentation. Notably, for UDA Semantic Segmentation on SYNTHIA, PCL surpasses the sophisticated CPSL-D by $>\!2\%$ in terms of mean IoU with a much lower training cost (PCL: 1*3090, 5 days v.s. CPSL-D: 4*V100, 11 days). Code is available at https://github.com/ljjcoder/Probabilistic-Contrastive-Learning.

\end{abstract} 

\section{Introduction}
\label{sec:intro}

\begin{figure}
	\centering
	\includegraphics[width=0.93\linewidth]{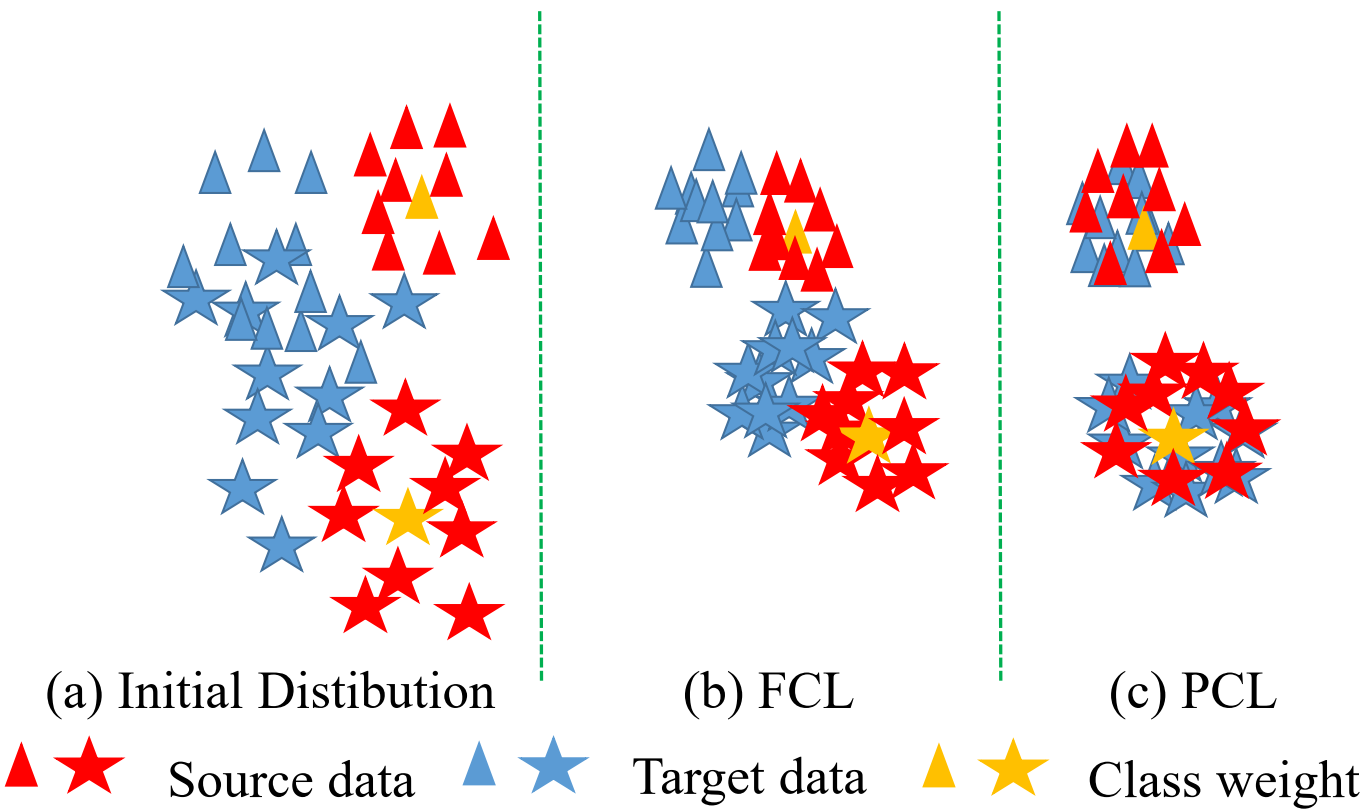}
	\caption{Feature Contrastive Learning (\textbf{FSL}) \vs Probabilistic Contrastive Learning (\textbf{PCL}). With PCL, the features on target domain can be clustered around the corresponding class weights.}
	\label{figure:motivation}
\end{figure}

Deep learning models are usually trained on a specific dataset (source domain) and perform well on similar dataset. However, when these models are applied to data from a different domain (target domain), their performance often degrades significantly. As illustrated in Figure~\ref{figure:motivation}(a), this is mainly because there exists domain shift or dataset bias between the source domain and the target domain. Domain adaptation~\cite{yan2017mind,na2021fixbi} offers a solution to this problem by allowing a model trained on a labeled source domain to adapt to an unlabeled or sparsely labeled target domain.
%
%
%

\begin{figure}
	\centering
	\includegraphics[width=0.95\linewidth]{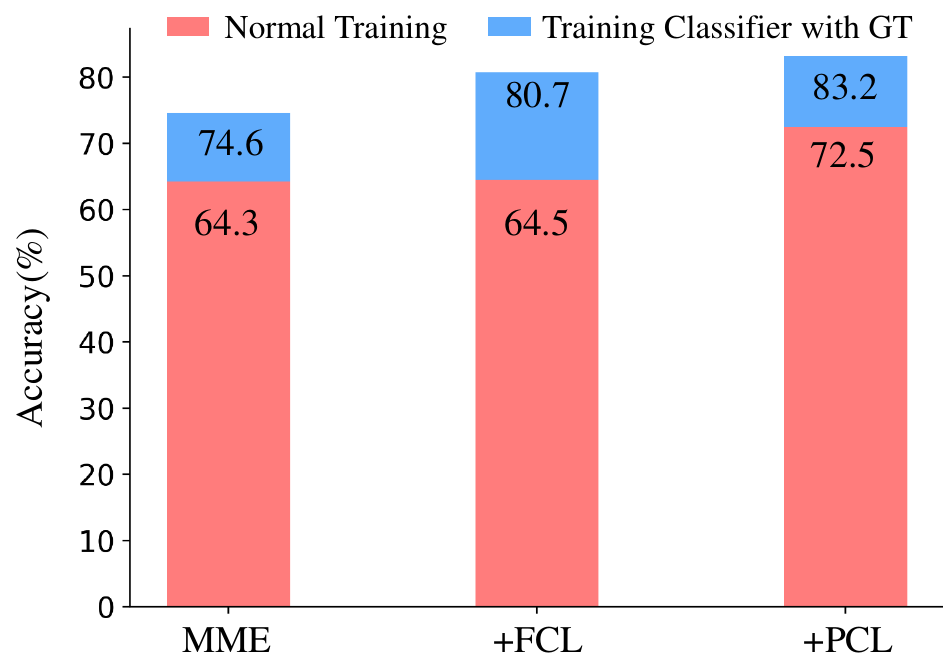}
	\caption{An explorative study under the SSDA setting on DomainNet (R$\rightarrow$S) with 3-shot and ResNet34.
	We use MME as a baseline model.}
	\label{figure:performance}
\end{figure}

For many visual tasks, the feature discriminability is the basis to obtain satisfying performance.
However, in domain adaptation, the learned features for each class on target domain are usually diffuse rather than discriminative as illustrated in Figure~\ref{figure:motivation}(a) since target domain lacks the ground-truth labels. 
Fortunately, contrastive learning is proposed to learn semantically similar features in a self-supervised manner~\cite{chen2020simple,khosla2020supervised}.
Inspired by its great success for representation learning, we hope to perform the standard contrastive learning (features+$\ell_{2}$ normalization) to assist feature extraction on unlabeled target domain.
However, we find that naively applying the standard contrastive learning in domain adaptation only brings very limited improvement (\eg, $64.3\% \rightarrow 64.5\%$ as shown in Figure~\ref{figure:performance}). 
A natural question arises and motivates this work: \emph{Why does contrastive learning perform poorly in domain adaptation?}
In the following, we first analyze the possible reasons and then propose a simple but powerful solution to this question.

For recognition tasks to achieve good performance, the learned features are not only required to be discriminative themselves, but also should be close to the class weights (\ie, the weights of the last fully connected layer). However, standard  contrastive learning usually typically utilizes features before the classifier to calculate the loss (here we term this method as Feature Contrastive Learning (FCL)). This approach does not involve the crucial class weights during optimization. While FCL can significantly enhance feature discriminability in the target domain, the issue of domain shift may still cause the features to deviate from the corresponding class weights learned from source data, as depicted in Figure~\ref{figure:motivation}(b). We validate this point using an experimental study. Specifically, we freeze the feature extractors and train a classifier borrowing the ground-truth labels on target domain (not available in the normal training). As shown in Figure~\ref{figure:performance}, the accuracy of FCL has increased by $6.1\%$ ($74.6\% \! \rightarrow \! 80.7\%$), but the actual accuracy has only increased by 0.2$\%$ ($64.3\% \! \rightarrow \! 64.5\%$). These findings lead us to propose that \emph{the deviation between features and class weights is the main reason that causes the poor performance of feature-based contrastive learning on domain adaptation}.
%
%
%

%
To deal with the deviation, we must introduce class weights information. Therefore, a na\"ive idea is to use the logits (\ie, the output of the last fully connected layer)  to calculate the contrastive loss. 
However, we experimentally find that simply introducing class weights information without explicit constraints cannot effectively alleviate the deviation problem.

This raises an important question: how can we develop a new type of contrastive loss that effectively alleviates the deviation between features and class weights? The key point is to dig out what kind of signal can effectively indicate that a feature vector is close to its corresponding class weights. For convenience, we define a set of class weights as $W=(\mathbf{w}_{1},\mathbf{w}_{2},\ldots,\mathbf{w}_{C})$, a feature vector as $\mathbf{f}_i$, and its classification probability as $\mathbf{p}_i$. Then the classification probability $p_{i,c}$ of $c$-th ($c\in\{1,2,\ldots, C\}$) class for feature $\mathbf{f}_i$ is 
\begin{small}
\begin{align*}
\small
p_{i,c} = \frac{ \exp(\mathbf{w}^{\top}_{c}\mathbf{f}_{i}) }{ \sum_{j \neq c} \exp( \mathbf{w}^{\top}_{j}\mathbf{f}_{i})+ \exp( \mathbf{w}^{\top}_{c}\mathbf{f}_{i})}.
\end{align*}
\end{small}
To ensure that feature vector $\mathbf{f}_i$ is closely aligned with its corresponding class weight $\mathbf{w}_{c}$, such that $\mathbf{w}^{\top}_c\mathbf{f}_i$ is large, $p_{i,c}$ will be close to 1 while $\{p_{i,j}\}_{j \neq c}$ approaches $0$.
This alignment suggests that as $\mathbf{f}_i$ gets closer to its corresponding class weights, the probability vector $\mathbf{p}_i$ will increasingly resemble a one-hot vector. 

To leverage this observation, we propose a novel contrastive learning framework, Probabilistic Contrastive Learning (PCL). 
PCL is different from standard methods by substituting features with probabilities and removing the $\ell_{2}$ normalization. These two straightforward yet impactful modifications enable PCL to impose a constraint on the probability vectors, guiding them towards a one-hot vector. This approach significantly mitigates deviation issues.
PCL is proposed as a simple but powerful method on domain adaptation, which is easily adaptable across various tasks and seamlessly integrable with different methodologies. Remarkably, PCL surpasses many complex alternatives, such as the meta-optimization in MetaAlign~\cite{wei2021metaalign} and prototypical+triplet loss in ECACL-P~\cite{li2021ecacl}.
Our main contributions are summarized as follows:
\begin{itemize}
    \item[1)] To the best of our knowledge, this is the first work to clearly point out that the problem of feature deviation from class weights is a core reason for the poor performance of standard FCL in domain adaptation tasks.
    \item[2)] Based on our analysis, we propose a new self-supervised paradigm called Probabilistic Contrastive Learning (PCL) for domain adaptation, which is simple in implementation and powerful in the generalization to different settings and various methods.
    \item[3)] Extensive experiments demonstrate that PCL can bring consistent performance improvements on different settings and various methods for domain adaptation.
\end{itemize}
%

\section{Related Work}

\subsection{Contrastive Representation Learning}
Contrastive learning is a mainstream representation learning method, which aims to learn a compact and transferable feature.
Currently,  extensive works~\cite{chen2020simple,dwibedi2021little,khosla2020supervised,khosla2020supervised} have demonstrated the effectiveness of this technique in a variety of vision tasks.
%
Some of these works consider conceiving elaborate model architectures to improve performance, such as memory bank~\cite{khosla2020supervised} or projection head~\cite{chen2020simple}.
%
%
%
%
However, these methods usually ignore class information, leading to the false negative problem.
%
%
Therefore, another part of the work focuses on how to select samples to alleviate the problem of false negative samples. For example, SFCL~\cite{khosla2020supervised} uses label information to eliminate wrong negative samples, effectively improving the performance. 
TCL~\cite{singh2021semi} extends this idea from instance-level to group-level and proposes group constrastive loss for semi-supervised action detection tasks.  
Different from previous methods, we argue that in domain adaptation tasks, the core bottleneck of contrastive learning is the deviation of features from class weights.  Based on this perspective, we design a simple yet efficient PCL, which does not have to rely on the techniques mentioned above, such as carefully designed positive and negative sample selection strategies and memory banks.

It is worth noting that, TCL enhances traditional architecture by extending the projection head to the classifier and softmax, the design that bears a resemblance to the form of PCL. However, in principle, PCL and TCL are still fundamentally different. Specifically, TCL is aimed at enhancing the selection strategy of samples, rather than addressing the issue of deviation between features and class weights. More importantly, it does not clarify the importance of probability, or even mention it.\footnote{In the paper of TCL, the defination of loss is based on the traditional contrastive paradigm (feature+$\ell_{2}$ normalization) and it does not specify the type of feature involved, whether it refers to the feature preceding the classifier, the logits, or the probabilities. The exact form it takes can only be known by examining the code.} It merely treats probability as a special feature, and naturally preserves $\ell_{2}$ normalization, that is, PCL-$\ell_{2}$. Therefore, whether in form or principle, TCL (PCL-$\ell_{2}$) is still a standard contrastive paradigm (features+$\ell_{2}$ normalization). Experiments show that TCL (PCL-$\ell_{2}$) are obviously inferior to PCL and we will discuss it in Sec.~\ref{app:LCL}.

\subsection{Domain Adaptation}
Domain adaptation mostly focuses on the recognition field (\eg, image classification, object detection, semantic segmentation) and aims to transfer the knowledge from a labeled source domain to an unlabeled target domain.
The literature can be roughly categorized into two categories.

The first category is to use domain alignment and domain invariant feature learning.
For example, \cite{yan2017mind} measures the domain similarity in terms of Maximum Mean Discrepancy (MMD), while \cite{peng2019moment} introduces the metrics based on second-order or higher-order statistics.
In addition, there are some methods~\cite{liu2019transferabletat,cui2020gradually} to learn domain-invariant features through adversarial training.

The second category is to learn discriminative representation using the pseudo-label technique~\cite{li2021ecacl,zhang2021prototypical}.
Particularly, in domain adaptive semantic segmentation, recent high-performing methods~\cite{zhang2021prototypical,li2022class} commonly use the distillation techniques.
Although distillation can greatly improve the accuracy,  it is time consuming and complex.

%

In this paper, we try to fully exploit the potential of contrastive learning for domain adaptation.
Our method can be well generalized to classification, detection, and segmentation tasks.
Particularly, in the domain adaptive semantic segmentation, PCL based on the non-distilled BAPA~\cite{liu2021bapa} can surpass CPSL-D~\cite{li2022class} which uses complex distillation techniques as well as special initialization strategies.
\section{Methods}
\label{method}
In this section, we first review feature contrastive learning (FCL), and then elaborate on our proposed probabilistic contrastive learning (PCL).
Generally speaking, we can split a model for classification, detection, and semantic segmentation into two parts: the encoder $E$ and the classifier $F$.
Here $F$ has the parameters $W=(\mathbf{w}_{1},...,\mathbf{w}_{C})$, where $C$ is the number of classes. Particularly, each vector in $W$ is equivalent to the embedding center of a class which we denote as class weights.


%
\subsection{Feature Contrastive Learning}
\label{method_fCL}
For domain adaptation~\cite{tzeng2014deep,long2017deep}, the source domain images already have clear supervision signals, and the self-supervised contrastive learning is not urgently required.
Thus, we only calculate the contrastive loss ({\textit{a.k.a.}}, InfoNCE~\cite{oord2018representation}) for the target domain data.
Specifically, let $\mathcal{B} = \{\left(x_{i},\tilde{x}_{i}\right)\}_{i=1}^{N}$ be a batch of data pairs sampled from target domain, where $N$ is the batch size, and $x_{i}$ and $\tilde{x}_{i}$ are two random transformations of a sample.
Then, we use $E$ to extract the features, and get $\mathcal{F}=\{(\mathbf{f}_{i}, \tilde{\mathbf{f}}_{i})\}_{i=1}^{N}$.
For a query feature $\mathbf{f}_{i}$, the feature $\tilde{\mathbf{f}}_{i}$ is the positive and all other samples are regarded as the negative.
Then the InfoNCE loss has the following form:

\begin{small}
\begin{align}
\small
	\ell_{\mathbf{f}_{i}} = \!
	- \!
	\log \! \frac{ \exp(sg(\mathbf{f}_{i}{})^{\top} g(\tilde{\mathbf{f}}_{i})) }{ \sum\limits_{j \neq i} \! \exp( sg(\mathbf{f}_{i}{})^{\top} \! g(\mathbf{f}_{j})) + \sum\limits_{k} \! \exp( sg(\mathbf{f}_{i}{})^{\top} \! g(\tilde{\mathbf{f}}_{k}))},
	\label{eq:FCL}
\end{align}
\end{small}%
where $g(\mathbf{f})=\frac{\mathbf{f}}{||\mathbf{f}||_2}$ is a standard $\ell_2$ normalization operation, and $s$ is the scaling factor. 


From Eq~(\ref{eq:FCL}), we can observe that there is no class weight information involved in $\ell_{\mathbf{f}_{i}}$.
As a result, in the optimization process of contrastive loss, it is hardly possible to constrain the features to locate around the class weights.

\subsection{A Na\"ive Solution}
Now, a natural question is: is it possible to effectively alleviate the deviation problem as long as the class weight information is introduced?
%
We choose logits (\ie, the class scores output by the classifier $F$) to calculate the contrastive loss as a naive method to introduce class weights information, and refer to it as Logits Contrastive Learning (LCL).
However, this approach does not explicitly cluster the features to be close to the class weights and thus does not work experimentally.
It shows that simply introducing class weight information without explicit constraints cannot achieve the goal.
%
Therefore, we need to design a new type of contrastive loss to explicitly reduce the deviation between the features and class weights.
More discussion about LCL will be provided in Section~\ref{app:LCL}.





\subsection{Probabilistic Contrastive Learning}
\label{method_PCL}
The generalization ability of InfoNCE~\cite{oord2018representation} has been fully verified in previous literature.
In this work, instead of designing a totally different loss function, we focus on constructing a new input $\mathbf{f}^{\prime}_{i}$ to calculate the contrastive loss for the sake of making the feature $\mathbf{f}_{i}$ close to class weights.
Formally, the loss about the new input $\mathbf{f}^{\prime}_{i}$ can be written as:
\begin{small}
\begin{align}
	\ell_{\mathbf{f}^{\prime}_{i}} =
	-
	\log \frac{ \exp(s{\mathbf{f}^{\prime}_{i}}^{\top} \tilde{\mathbf{f}^{\prime}}_{i}) }{ \sum_{j \neq i} \exp( s{\mathbf{f}^{\prime}_{i}}^{\top} \mathbf{f}^{\prime}_{j})+  \sum_{k} \exp( s{\mathbf{f}^{\prime}_{i}}^{\top} \tilde{\mathbf{f}^{\prime}}_{k})}.
	\label{eq:FCL_temp}
\end{align}
\end{small}%
Then our goal is to design a suitable $\mathbf{f}^{\prime}_{i}$ so that
\emph{the smaller $\ell_{\mathbf{f}^{\prime}_{i}}$ is, the closer $\mathbf{f}_{i}$ is to the class weights.}

From Eq~(\ref{eq:FCL_temp}), a smaller $\ell_{\mathbf{f}^{\prime}_{i}}$ means a larger ${\mathbf{f}^{\prime}_{i}}^{\top} \tilde{\mathbf{f}^{\prime}}_{i}$.
Thus the above problem can be roughly simplified to \emph{the larger ${\mathbf{f}^{\prime}_{i}}^{\top} \tilde{\mathbf{f}^{\prime}}_{i}$ is, the closer $\mathbf{f}_{i}$ is to the class weights.}
On the other hand, as elaborated in Section~\ref{sec:intro}, if $\mathbf{f}_{i}$ is close to the class weight, the corresponding probability ${\mathbf{p}_{i}}$ is approximate to the one-hot form:
\begin{small}
\begin{align}
	{\mathbf{p}_{i}} =(0,..,1,..,0).
	\label{eq:peak-distribution}
\end{align}
\end{small}%
Therefore, our goal can be reformulated as how to design a suitable $\mathbf{f}^{\prime}_{i}$ so that
\emph{the larger ${\mathbf{f}^{\prime}_{i}}^{\top} \tilde{\mathbf{f}^{\prime}}_{i}$ is, the closer $\mathbf{p}_{i}$ is to the one-hot form.} 



\begin{figure}
	\centering
	\includegraphics[width=0.93\linewidth]{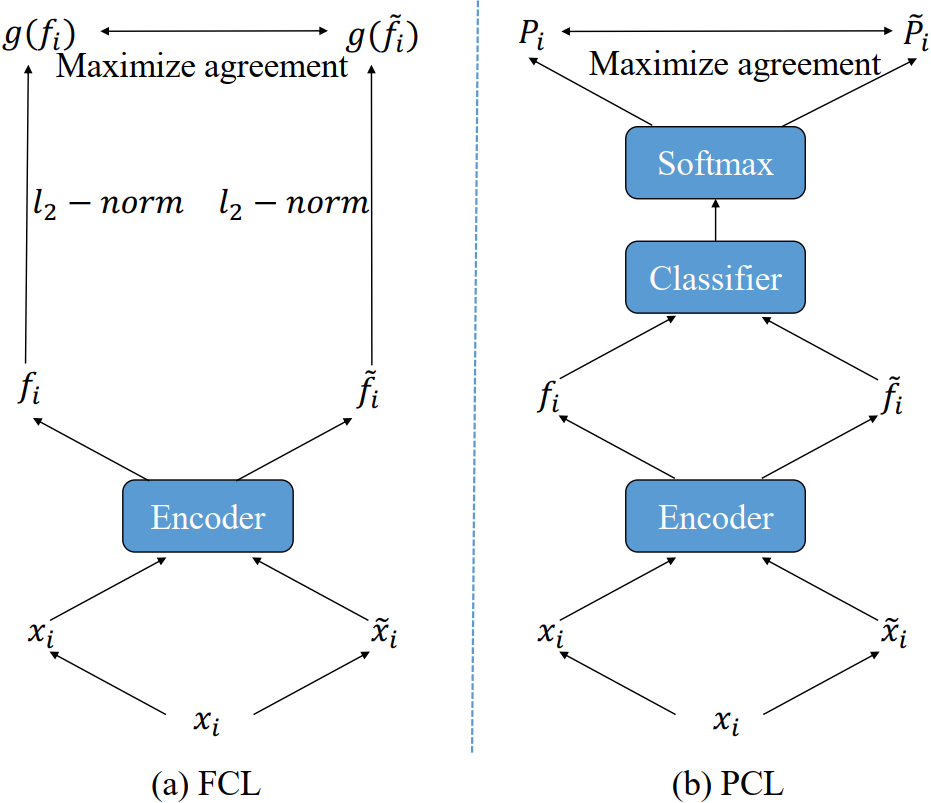}
	\caption{Framework of FCL and PCL. Different from FCL, PCL uses the output of softmax to perform contrastive learning and removes the $\ell_{2}$ normalization.}
	\label{figure:Framework}
\end{figure}

Fortunately, we found that the probability $\mathbf{p}_{i}$ itself can meet such a requirement.
Here we explain the mathematical details.
Note that  $\mathbf{p}_{i}=(p_{i,1},...,p_{i,C})$ and $\tilde{\mathbf{p}}_{i}=(\tilde{p}_{i,1},...,\tilde{p}_{i,C})$ are both the probability distributions. Then we have
\begin{small}
\begin{align}
	{0\leq p_{i,c} \leq 1}, {0\leq \tilde{p}_{i,c} \leq 1},   {\forall}c \in \{1,...,C\}.
	\label{eq:prob_feature}
\end{align}
\end{small}%
In addition, the $\ell_{1}$-norm of $\mathbf{p}_{i}$ and $\tilde{\mathbf{p}}_{i}$ equals one, \textit{i.e.}, $||\mathbf{p}_{i}||_1=\sum_{c} p_{i,c}=1$ and $||\tilde{\mathbf{p}}_{i}||_1=\sum_{c} \tilde{p}_{i,c}=1$. Obviously, we have
\begin{small}
\begin{align}
	\mathbf{p}_{i}{}^{\top} \tilde{\mathbf{p}}_{i}=\sum_{c}p_{i,c}\tilde{p}_{i,c}\leq 1.
	\label{eq:prob_dot}
\end{align}
\end{small}%
The equality is held if and only if $\mathbf{p}_{i} =\tilde{\mathbf{p}}_{i}$ and both of them have a one-hot form as in Eq~(\ref{eq:peak-distribution}).
In other words, in order to maximize $\mathbf{p}_{i}^{\top} \tilde{\mathbf{p}}_{i}$, the $\mathbf{p}_{i}$ and $\tilde{\mathbf{p}}_{i}$ need to satisfy the one-hot form at the same time.
Therefore, $\mathbf{p}_{i}$ can be served as the new input $\mathbf{f}^{\prime}_{i}$ in Eq~(\ref{eq:FCL_temp}).

Importantly, from the above derivation process, we can see that \emph{the property that the $\ell_{1}$-norm of probability equals one is very important}. This property guarantees that the maximum value of $\mathbf{p}_{i}^{\top} \tilde{\mathbf{p}}_{i}$ can only be reached when $\mathbf{p}_{i}$ and $\tilde{\mathbf{p}}_{i}$ satisfy the one-hot form at the same time. Evidently, we cannot perform $\ell_{2}$ normalization operation on probabilities like the traditional FCL.
Finally, our new contrastive loss is defined by
\begin{small}
\begin{align}
	\ell_{\mathbf{p}_{i}} =
	-
	\log \frac{ \exp(s{\mathbf{p}_{i}}^{\top} \tilde{\mathbf{p}}_{i}) }{ \sum_{j \neq i} \exp( s{\mathbf{p}_{i}}^{\top} \mathbf{p}_{j})+  \sum_{k} \exp( s{\mathbf{p}_{i}}^{\top} \tilde{\mathbf{p}}_{k})}.
	\label{eq:D-InfoNCE}
\end{align}
\end{small}

Figure~\ref{figure:Framework} gives an intuitive comparison between FCL and PCL and we can see two main differences.
First, Eq~(\ref{eq:D-InfoNCE}) uses the probability $\mathbf{p}_{i}$ instead of the extracted features $\mathbf{f}_{i}$.
Second, Eq~(\ref{eq:D-InfoNCE}) removes the $\ell_{2}$ normalization $g$.
It is worth emphasizing that, the rationale behind PCL is the core value of this work, which leads to a convenient implementation.
Benefiting from the conciseness, PCL can well generalized to different settings and various methods.

\section{Discussion}

%
In this work, we re-examine contrastive learning in domain adaptation from a new perspective, not just based on the broad perspective of ``contrastive learning can improve the generalization of features or effectively utilize unlabeled data"~\cite{singh2021clda,singh2021semi}. Specifically,  we argue that in domain adaptation tasks, the core reason for the poor performance of contrastive learning is that traditional FCL cannot effectively narrow the distance between features and class weights.
Based on the above insights, we propose the PCL  and surprisingly find that only employing two simple operations (using probabilities and removing $\ell_{2}$ normalization), without any other techniques, can greatly alleviate the deviation problem.
Few works have examined the challenges of contrastive learning in domain adaptation from this perspective, making our analysis and identification of this shortcoming a significant and novel contribution of this paper.
%

On the other hand, PCL bears similarities to some existing works due to its simplicity. For example, PCL can easily be thought of as a special projection head~\cite{chen2020simple}; it can also be viewed as a special entropy minimization loss ~\cite{Chen_2019_ICCV,zhong2021neighborhood}. This similarity raises the suspicion that the reason why PCL is effective is just the application of these techniques rather than the novel points we claim. In addition, there are many works~\cite{dwibedi2021little,khosla2020supervised} that have greatly improved FCL by mitigating the false negative sample problem. It easily makes us wonder whether PCL is still necessary when false negative samples are mitigated.

In this section, we demonstrate through comprehensive quantitative comparisons that our insights and proposed PCL are the key points in resolving deviation issues and enhancing domain adaptation performance. In the quantitative comparison, we use the typical semi-supervised domain adaptation (SSDA) setting on DomainNet~\cite{peng2019moment} with 3-shot and ResNet34 as our benchmark. In particular, we choose MME~\cite{saito2019semi} as the baseline model. To ensure fairness, we also keep the training strategy and parameters completely consistent. The comparisons are organized as follows:
In Sec.~\ref{app:PCL_FCL}, we verify whether PCL is better than FCL. In Sec.~\ref{app:LCL}, Sec.~\ref{app:cos_mse}, Sec.~\ref{app:PCL_EM}, we compare a series of techniques similar to PCL. In Sec.~\ref{sec:False Negative}, we discuss the false negative problems and deviation problems. Finally, in Sec.~\ref{sec:vis}, we show the visualization results. We believe these analyses can also provide some useful insights for other visual tasks~\cite{mohri2019agnostic,li2021learning}.





%
%

%
\subsection{PCL v.s. FCL}
\label{app:PCL_FCL}

In this part, we compare contrastive learning based on features (FCL) and probabilities (PCL), and present the results in Table~\ref{tab:different_feature_position}. 
It can be seen that PCL can greatly improve the gain of FCL (FCL: 1.8$\%$ v.s. PCL: 7.4$\%$). 
%

\begin{table}
\centering

\resizebox{\linewidth}{!}{
\begin{tabular}{c|ccccccc|c}
\specialrule{.1em}{.05em}{.05em}
 Method & R$\rightarrow$C & R$\rightarrow$P & P$\rightarrow$C & C$\rightarrow$S & S$\rightarrow$P & R$\rightarrow$S & P$\rightarrow$R &Mean  \\
\hline
   Baseline & 71.4 & 70.0 & 72.6 & 62.7 & 68.2 & 64.3 & 77.9 & 69.5 \\ 
\hline
   + FCL & 72.5 & 71.6 & 73.1 & 66.4 & 70.2 & 64.5 & 80.8 & 71.3 \\ 
   + NTCL & 72.9 & 71.3 & 73.3 & 66.3 & 71.3 & 67.1 & 80.5 & 71.7 \\
   + LCL & 72.8 & 70.6 & 72.5 & 66.4 & 70.5 & 64.5 & 81.3 & 71.2 \\
   + PCL-$\ell_{2}$ & 75.1 & 74.4 & 76.2 & 70.3 & 73.5 & 69.9 & 82.5 & 74.6 \\ 
   + \textbf{Our PCL} & \textbf{78.1} & \textbf{76.5} & \textbf{78.6} & \textbf{72.5} & \textbf{75.6} & \textbf{72.5} & \textbf{84.6}  & \textbf{76.9} \\ 
\hline
\specialrule{.1em}{.05em}{.05em}
\end{tabular}}
\vspace{0.3 mm}
\caption{Classification accuracy (\%) of different features on DomainNet under the setting of $3$-shot and Resnet34.}
\label{tab:different_feature_position}

\end{table}

\subsection{PCL v.s. FCL with Projection Head}
\label{app:LCL}
Projection head~\cite{chen2020simple} is a very useful technique that changes the paradigm from feature+$\ell_{2}$ normalization to projection head feature+$\ell_{2}$ normalization. Inspired by this, CLDA~\cite{singh2021clda} uses the classifier as the projection head to design two contrastive learning losses and achieve better performance. PCL is similar in form to using the classifier as the projection head.
In this section, we mainly verify whether the performance gain of our PCL comes from the application of the projection head. To this end, we designed the following three types of projection heads:

(1) Following the SimCLR~\cite{chen2020simple}, we introduce an additional nonlinear transformation (NT) on the feature. We call it NT-Based Contrastive Learning (NTCL).

(2) We directly use the classifier as the projection head and named it Logits Contrastive Learning (LCL) to introduce class weight information. 

(3) We further generalize the projection head to classifier+softmax. For this setting, the contrastive loss paradigm becomes classifier+softmax with $\ell_{2}$ normalization. Essentially, it has only one more $\ell_{2}$ normalization than our PCL, and thus we denote it PCL-$\ell_{2}$. Like LCL, PCL-$\ell_{2}$ is also a way to introduce class weight information. However, it is worth noting that PCL-$\ell_{2}$ is not a natural extension of the projection head technique, since current mainstream contrastive learning methods ~\cite{chen2020simple,dwibedi2021little,chen2021empirical,chen2021exploring} do not include softmax in the projection head. 
The purpose of this design is to verify such a question:  \textbf{when probability is used as a special feature, can the traditional contrastive learning paradigm (feature + $\ell_{2}$ normalization) be as effective as PCL?} 
%

Table~\ref{tab:different_feature_position} gives the experimental results and we obtain the following observations.
\textbf{First}, the above three projection heads all get lower performance than PCL, which indicates that the key reason for the gain of PCL is not from the use of the projection head.
\textbf{Second}, both LCL and PCL-$\ell_{2}$ are inferior to PCL, which shows that simply introducing class weight information cannot effectively enforce features to gather around class weights. It also means that to reach the goal, the loss function needs to be carefully designed. 
\textbf{Third}, the results experimentally verify the importance of core motivation. Without realizing the problem of features deviating from class weights, we cannot break out of the standard paradigm and induce PCL. 
Because we have neither reason to use probability nor reason to abandon the widely used $\ell_{2}$ normalization.
Even if, like TCL~\cite{singh2021semi}, happens to extend the projection head to softmax, it still has no enough motivation to remove the $\ell_{2}$ normalization that is widely used in contrastive learning. 
In this work, however, our core motivation is exactly that the deviation between the features and class weights is a key factor affecting contrastive learning performance. Based on this, we naturally induce the concise form of PCL (in Sec.~\ref{method_PCL}). 


\subsection{Cosine distance v.s. MSE distance}
\label{app:cos_mse}
In domain adaptation, there has been a work~\cite{french2017self} that exploits the similarity of the prediction space for consistency constraints, although it does not use the form of contrastive learning. Intuitively, PCL seems to just apply this idea to the contrastive learning loss (transfer feature space to prediction space). Therefore, we need to answer an important question:  \textbf{is PCL effective only because of the consistency constraint in the prediction space?}

From the previous analysis, PCL naturally requires the use of probabilistic cosine similarity to narrow the distance between features and class weights, rather than just requiring the consistency of the output space. To verify it, we replace the inner product in PCL with the MSE used in ~\cite{french2017self} to get PCL-MSE. Table~\ref{tab:SFCL_BCE} gives the results. It can be seen that PCL is better than PCL-MSE. This is because mse can only make the probability similar but not make the probability appear in the one-hot form. This again proves the importance of the motivation of PCL, because this motivation ensures that PCL must use the cosine distance but not the MSE distance.


\subsection{PCL v.s. Entropy Minimization}
\label{app:PCL_EM}
\begin{table}

\centering
\resizebox{\linewidth}{!}{
\begin{tabular}{c|ccccccc|c}
\specialrule{.1em}{.05em}{.05em}
 Method & R$\rightarrow$C & R$\rightarrow$P & P$\rightarrow$C & C$\rightarrow$S & S$\rightarrow$P & R$\rightarrow$S & P$\rightarrow$R &Mean  \\
\hline
   Baseline & 71.4 & 70.0 & 72.6 & 62.7 & 68.2 & 64.3 & 77.9 & 69.5 \\
\hline
   + FCL & 72.5 & 71.6 & 73.1 & 66.4 & 70.2 & 64.5 & 80.8 & 71.3 \\
   + SFCL & 72.7 & 72.4 & 73.2 & 66.5 & 70.8 & 65.5 & 81.2 & 71.8 \\
   + BCE & 73.3 & 73.0 & 74.7 & 66.7 & 71.9 & 67.5 & 80.4 & 72.5 \\
   + maxsqures & 73.9 & 72.3 & 74.0 & 66.3 & 71.3 & 67.6 & 80.3 & 72.2 \\
   + FCL+maxsqures & 72.9 & 72.0 & 73.1 & 66.7 & 70.7 & 66.8 & 80.7 & 71.8 \\
   + FCL+BCE & 73.9 & 73.1 & 74.1 & 66.5 & 71.5 & 67.3 & 81.4 & 72.5 \\
   + Our PCL-MSE & 76.6 & 75.3 & 76.2 & 69.7 & 74.2 & 70.5 & 83.8 & 75.1 \\
   + \textbf{PCL} & \textbf{78.1} & \textbf{76.5} & \textbf{78.6} & \textbf{72.5} & \textbf{75.6} & \textbf{72.5} & \textbf{84.6}  & \textbf{76.9} \\
\hline
\specialrule{.1em}{.05em}{.05em}
\end{tabular}}
\vspace{0.3 mm}
\caption{Classification accuracy (\%) of different FCL improvement methods on DomainNet under the setting of $3$-shot and Resnet34.}
\label{tab:SFCL_BCE}
\end{table}
PCL forces probabilities to approximate one-hot form, thereby reducing the entropy of predictions. Therefore, PCL can also be regarded as an entropy minimization loss.
Naturally, it also raises an important question:  \textbf{can we achieve the similar performance by using other classic entropy minimization losses, such as maxsqures loss~\cite{Chen_2019_ICCV} and binary cross entropy loss (BCE)~\cite{zhong2021neighborhood}?}   



\begin{figure}
    \centering

        \centering
     \includegraphics[width=1.0\columnwidth]{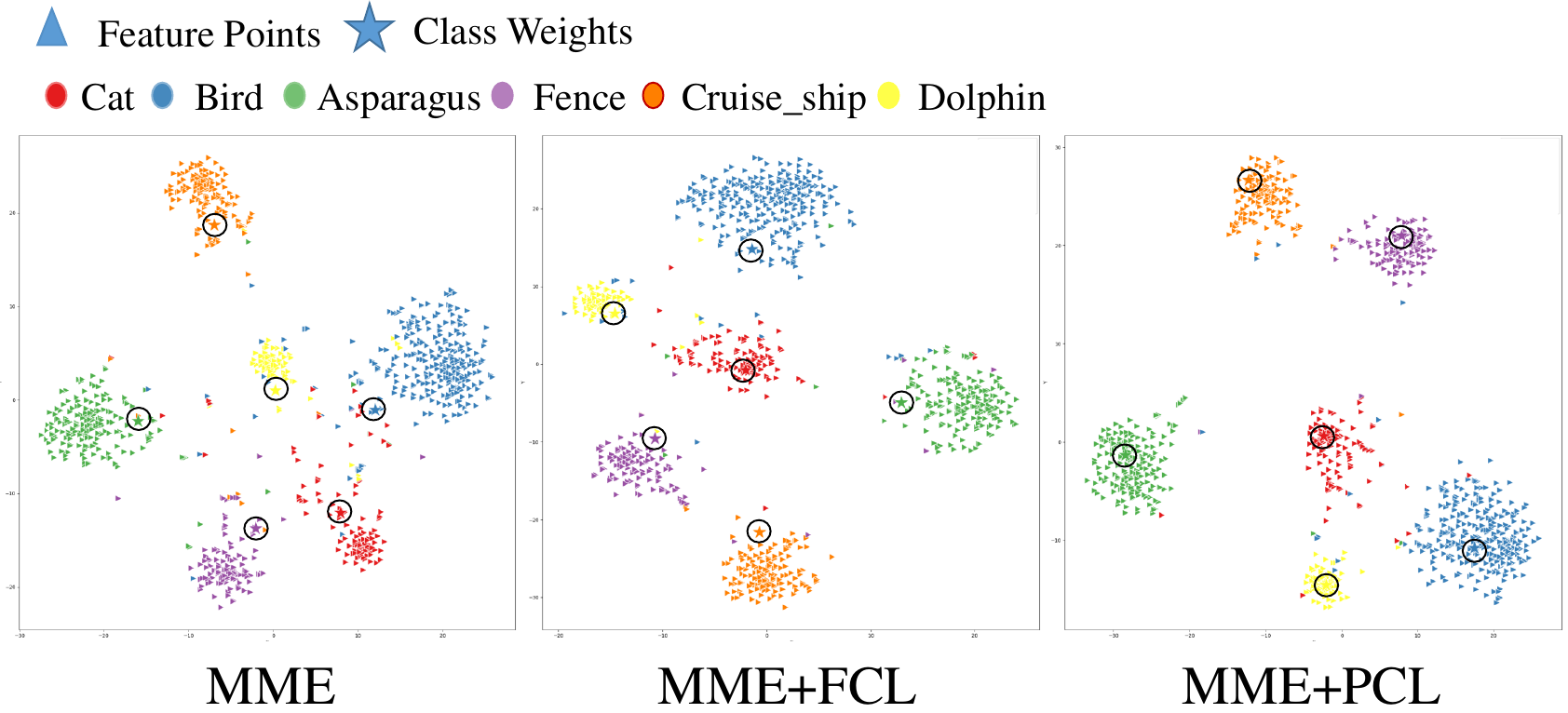}

     \caption{The t-SNE visualization of learned features. Best viewed in color.}
     \label{fig:tsne}

\end{figure}
Table~\ref{tab:SFCL_BCE} gives the results and we have the following observations.

First, like PCL, the optimization goals of maxsquares loss and BCE loss can force the probability to approach the one-hot form. Therefore, they can indeed bring gains based on the baseline (\eg, maxsquare: $69.5\% \rightarrow 72.2\%$, BCE: $69.5\% \rightarrow 72.5\%$).

Second, compared to PCL, the gains of maxsquares loss and BCE loss are very limited, which reflects the fact that maintaining the InfoNCE format is critical to the success of PCL. As we emphasized in Sec.\ref{sec:intro}, for recognition tasks, we not only require the features to be close to the class weights, but also require the features themselves to be compact enough. However, for BCE loss and maxSquares loss, although they can also make the probability approach the one-hot form, they use a pair-wise form of loss to constrain the sample (maxsqure loss even discards negative samples and different data augmentations), which is far less effective than the InfoNCE form in learning compact feature representations ~\cite{wang2021understanding}.


\subsection{PCL v.s. SFCL}
\label{sec:False Negative}
In original contrastive learning loss, there may be some negative samples that belong to the same category as the query sample, which are called false negative samples.
Many methods~\cite{dwibedi2021little,khosla2020supervised} point out that false negative samples are harmful.
In this section, we try to answer another important question: \textbf{whether can we make FCL work well by reducing the false negative samples without designing PCL? }

%
%
%

An appropriate way to address the false negative samples problem is to use supervised feature contrastive learning (SFCL)~\cite{khosla2020supervised}. 
Table~\ref{tab:SFCL_BCE} shows the experimental comparison.
It can be seen that SFCL can indeed improve the performance of FCL.
However, compared with PCL, SFCL has a very limited improvement over FCL. Specifically, SFCL can learn better feature representations by alleviating the false negative problem, but it cannot solve the problem of the deviation between the features and class weights. The experimental results reveal that the deviation problem is more critical than the false negative samples for domain adaptation. 

\subsection{Visualization Analysis}
\label{sec:vis}
Figure~\ref{fig:tsne} shows the relationship between the unlabeled features and the class weights for the three methods, including MME, MME+FCL, and MME+PCL.
Firstly, compared with MME, MME+FCL produces more compact feature clusters for the same category and more separate feature distributions for different categories. However, the learned class weights deviate from the feature centers for both MME+FCL and MME.
Secondly, the class weights of MME+PCL are much closer to the feature centers than MME+FCL. It demonstrates that PCL is significantly effective in enforcing the features close to the class weights.


\section{Experiments}
In this section, we will verify the validity of the PCL on five different tasks. In order to ensure fairness, in each task, we strictly follow the baseline experimental settings and only add additional PCL loss. 

\subsection{UDA Semantic Segmentation}
\label{app:rudass}
\begin{table}

	\centering
	\renewcommand{\arraystretch}{1.2}
	\setlength{\tabcolsep}{8pt}
	\renewcommand{\arraystretch}{1.2}
	\resizebox{0.45\textwidth}{!}{
	\begin{tabular}{cccc}
		
		\toprule
		\multirow{2}*{Methods}  				& \multicolumn{1}{c}{GTA5}	&\multicolumn{2}{c}{SYNTHIA} 		\\
		\cline{2-4}
		\multicolumn{1}{c}{ } &mIoU (\%) &mIoU-13 (\%) &mIoU-16 (\%) \\
		\hline

		ProDA ( CVPR'21 )~\cite{zhang2021prototypical}& 53.7& -& -				\\
            CPSL( CVPR'22 )~\cite{li2022class}& 55.7& 61.7& 54.4				\\
            BAPA ( ICCV'21 )~\cite{liu2021bapa}	& 57.4& 61.2& 53.3\\
  		\hline
            ProDA-D ( CVPR'21 )~\cite{zhang2021prototypical} & 57.5& 62.0& 55.5				\\
            ProDA-D+CaCo ( CVPR'22 )~\cite{huang2022category}& 58.0& -& -				\\
            CPSL-D ( CVPR'22 )~\cite{li2022class}& \textbf{60.8}& 65.3& 57.9				\\
    		\hline
		BAPA$^*$					& 57.7& 60.1& 53.3				\\
		+ \textbf{Our PCL}					& 60.7& \textbf{68.2}& \textbf{60.3}	\\
		\bottomrule	
	\end{tabular}}
 \vspace{0.6 mm}
 	\caption{Result on UDA Semantic Segmentation. -D means to use an additional two-step distillation technique. ${*}$ means our reimplementation. }
	\label{table:BAPA_ours}
\end{table}

\noindent\textbf{\textit{Setup}} We evaluate our method on two standard UDA semantic segmentation tasks: GTA5~\cite{richter2016playing}$\rightarrow$Cityscapes~\cite{cordts2016cityscapes} and SYNTHIA~\cite{ros2016synthia}$\rightarrow$Cityscapes.

The current SOTA methods generally adopt the distillation technique for post-processing. It makes the training process very complicated and requires some special training strategies. Therefore, here we divide these methods into simple non-distilled methods and complex distillation methods. In particular, we take the non-distilled BAPA~\cite{liu2021bapa} with ResNet-101~\cite{ResNet1} as our baseline due to its simplicity and efficiency. For the hyperparameter in PCL, we set $s=20$ in all experiments. 

Table~\ref{table:BAPA_ours} gives the results. First, our method can achieve very significant gains on the baseline and outperforms all non-distilled methods by a large margin on SYNTHIA (6.5$\%$ for mIoU-13 and 5.9$\%$ of mIoU-16). Second, even compared to distillation-based methods, our method has only slightly lower performance than CPSL-D on GTA5 and outperforms CPSL-D by more than 2$\%$ on SYNTHIA. Notably, the training cost of our method is much lower than  CPSL-D (PCL: 1*3090, 5 days v.s. CPSL-D: 4*V100, 11 days). 

\subsection{UDA  Detection}
\label{app:DAOD}
\noindent\textbf{\textit{Setup}} We conduct an experiment on SIM10k~\cite{SIM10K} $\rightarrow$ Cityscapes~\cite{cordts2016cityscapes} scenes to verify effective of our PCL for the object detection task.
In particular, we choose the RPA~\cite{zhang2021rpn} with Vgg16~\cite{simonyan2014very} as the baseline. 
We add PCL to the classification head to improve the classification results of the RPA model. For the hyperparameter in PCL, we set $s=20$ in the experiment. 

Table~\ref{table:sim10k_city} gives the results. It can be seen that PCL can significantly improve the performance of the RPA model. This proves the effectiveness of PCL on the UDA detection task.

\begin{table}

\centering
\renewcommand{\arraystretch}{1.0}
\resizebox{\linewidth}{!}{
\begin{tabular}{cc|cc}
\specialrule{.1em}{.05em}{.05em}
					\hline
					Method & AP & Method & AP \\
					\hline
					MeGA-CDA  ( CVPR'21 )~\cite{VS_2021_CVPR}  & 44.8 & RPA ( CVPR'21 )$^{*}$~\cite{zhang2021rpn} & 45.3 \\
					  UMT ( CVPR'21 )~\cite{deng2021unbiased} & 43.1    &  + \textbf{Our PCL} &  \textbf{47.8} \\
					\hline
					\specialrule{.1em}{.05em}{.05em}
				\end{tabular}}
    \vspace{0.3 mm}
    \caption{Detection performance (\%) on UDA detection task. ${*}$ means our reimplementation.}

\label{table:sim10k_city}
\end{table}




\subsection{UDA Classification}
\label{app:ruda}

\noindent\textbf{\textit{Setup}} We evaluate our PCL in the following two standard benchmarks: \textbf{Office-Home}~\cite{Office-HOME} and \textbf{VisDA-2017}~\cite{VisDA2017}. 
 We take GVB-GD~\cite{cui2020gradually} with ResNet50~\cite{ResNet1} as our baseline. For the hyperparameter in PCL, we set $s=7$ in all experiments. 
\begin{table}

\centering
\renewcommand{\arraystretch}{1.1}
\resizebox{\linewidth}{!}{
\begin{tabular}{cc|cc}
					\hline
					Method & Acc & Method & Acc \\
					\hline
     SDAT  ( ICML'22 )~\cite{rangwani2022closer}  & 72.2 & SCDA ( ICCV'21 )~\cite{li2021semantic}  & 73.1 \\
     NWD ( CVPR'22 )~\cite{chen2022reusing} & 72.6 & HMA ( ICCV'23 )~\cite{zhou2023homeomorphism} & 73.2 \\
				    \hline
					 GVB* ( CVPR'20 )~\cite{cui2020gradually} & 70.3 &  GVB$^\dag$ &  73.5 \\
      			 + MetaAlign ( CVPR'21 )~\cite{wei2021metaalign} & 71.3 &  - &  -\\
					+ \textbf{Our PCL} &  \textbf{72.3}  &   + \textbf{Our PCL}  & \textbf{74.5} \\
					\hline
					\specialrule{.1em}{.05em}{.05em}
				\end{tabular}}
    \vspace{0.3 mm}
    \caption{Average performance (\%) of 12 UDA tasks on Office-Home. * means our reimplementation. }  
\label{tab:uda_officehome} 
\end{table}

\begin{table}

\centering
\renewcommand{\arraystretch}{1.1}
\resizebox{\linewidth}{!}{
\begin{tabular}{cc|cc}
\specialrule{.1em}{.05em}{.05em}
					\hline
					Method & Acc & Method & Acc \\
					\hline
					CST ( NeurIPS'21 )~\cite{liu2021cycle}  & 80.6 & SENTRY ( ICCV'21 )~\cite{prabhu2021sentry} & 76.7 \\
				    \hline
					 GVB* ( CVPR'21 )~\cite{cui2020gradually} & 75.0 &  GVB$^\dag$  &  80.4 \\
					+ \textbf{Our PCL} &  80.8 &   + \textbf{Our PCL}  & \textbf{82.5} \\
					\hline
					\specialrule{.1em}{.05em}{.05em}
				\end{tabular}}
    \vspace{0.3 mm}
    \caption{Classification accuracy (\%) of Synthetic$\rightarrow$Real on VisDA-2017 for UDAs. * means our reimplementation.}  
\label{tab:uda_visda}
\end{table}

Table~\ref{tab:uda_officehome} and Table~\ref{tab:uda_visda} give the results.
In particular, inspired by some previous semi-supervised domain adaptation methods~\cite{li2021cdac,li2021ecacl}, we add FixMatch~\cite{fixmatch} to GVB and get the stronger GVB$^\dag$. 
It can be seen that our method can bring considerable gains both on GVB and GVB$^\dag$. In particular, with GVB as the baseline, PCL outperforms MetaAlign by $1\%$, which involves more complicated operations than PCL. This further demonstrates the superiority of PCL.

\subsection{Semi-Supervised Domain Adaptation}
\label{app:rssda}

\begin{table}

	\centering
	\renewcommand{\arraystretch}{1.2}
	\setlength{\tabcolsep}{8pt}
	\renewcommand{\arraystretch}{1.2}
	\resizebox{0.45\textwidth}{!}{
	\begin{tabular}{cccc}
		
		\toprule
		\multirow{2}*{Methods}  				& \multicolumn{1}{c}{Office-Home}	&\multicolumn{2}{c}{DomainNet} 		\\
		\cline{2-4}
		\multicolumn{1}{c}{ } &3-shot  &1-shot &3-shot \\
		\hline

		CLDA ( NeurIPS'21 ) ~\cite{singh2021clda}& \textbf{75.5} & 71.9 & 75.3				\\
            \cdashline{1-4}[0.8pt/2pt]
            MME$^{*}$ ( ICCV'19 )~\cite{saito2019semi}	& 73.5 & 67.9 & 69.5\\
            + \textbf{Our PCL} & \textbf{75.5} & \textbf{73.5} & \textbf{76.9}				\\
  		\hline

            ECACL-P$^\dag$  ( ICCV'21 )~\cite{li2021ecacl} & -& 72.8 & 76.4 \\
            MCL$^\dag$ ( IJCAI'22 )~\cite{yan2022multi} & 77.1& 74.4 & 76.5 \\
            ProMM$^\dag$ ( IJCAI'23 )~\cite{huang2023semi}& 77.8& 76.1 & 77.4 \\
            \cdashline{1-4}[0.8pt/2pt]
            CDAC$^\dag$ ( CVPR'21 )~\cite{li2021cdac}	& 74.8 & 73.6& 76.0\\
            +SLA ( CVPR'23 )~\cite{yu2023semi} & 76.3 & 75.0& 76.9            
            \\
            \cdashline{1-4}[0.8pt/2pt]
            MME$^\dag$& 76.9& 72.9&76.1				\\
		+ \textbf{Our PCL}					& \textbf{78.1}& \textbf{75.1}& \textbf{78.2}	\\
		\bottomrule	
        \end{tabular}}
 \vspace{0.3 mm}
	\caption{Average performance (\%) on  \textit{DomainNet} (7 tasks) and \textit{Office-Home} (12 tasks). ${*}$ means our reimplementation.}
	\label{tab:ssda_domainnet_officehome}
\end{table}
\noindent\textbf{\textit{Setup}} We evaluate the effectiveness of our proposed approach on two SSDA benchmarks, \textit{i.e.},  \textit{DomainNet}~\cite{peng2019moment} and \textit{Office-Home}.
We choose MME~\cite{saito2019semi} with ResNet34~\cite{ResNet1} as our baseline model. In particular, inspired by CDAC~\cite{li2021cdac} and ECACL-P~\cite{li2021ecacl}, we add FixMatch~\cite{fixmatch} to MME to build a stronger MME$^\dag$. For the hyperparameter in PCL, we set $s=7$ in all experiments. 



Table~\ref{tab:ssda_domainnet_officehome} gives the results and we obtain the following observations:
\noindent1) PCL outperforms the methods without Fixmatch for most of settings. In particular, CLDA uses a classifier as the projection head for instance-level and class-level contrastive learning. Our PCL can defeat CLDA, although PCL is only equipped with instance contrastive learning.  The results indicate the superiority of PCL over the projection head.
\noindent2) Comparing with the methods using Fixmatch, PCL has obvious advantages in terms of simplicity and effectiveness. First of all, these methods, in addition to using Fixmatch, also carefully design many complex strategies to enhance performance. For example, ECACL-P designs the prototypical loss and triplet loss. 
Even so, without relying on Fixmatch, simple PCL can almost equal these complex methods. After equipping Fixmatch, PCL achieved clear advantages in all settings. 

		

\subsection{Semi-Supervised Learning}
\label{app:rssl}
\begin{table}
 
\centering
\renewcommand{\arraystretch}{1.1}
\resizebox{\linewidth}{!}{
\begin{tabular}{cc|cc}
\specialrule{.1em}{.05em}{.05em}
					\hline
					Method & 4-shot Acc & Method & 4-shot Acc \\
					\hline
					Freematch  ( ICLR'23 )~\cite{wang2022freematch}  & 62.02$\pm$0.42 & Softmatch ( ICLR'23 )~\cite{chen2023softmatch} & 62.90$\pm$0.77 \\					
				    \hline
					 FixMatch* ( NeurIPS'20 )~\cite{fixmatch} & 53.58$\pm$ 2.09 &  + \textbf{Our PCL} &  \textbf{57.62$\pm$2.52} \\
					CCSSL* ( CVPR'22 )~\cite{yang2022class} &  60.49$\pm$0.57  &   + \textbf{Our PCL}  & \textbf{62.95}$\pm$1.39 \\
					 FlexMatch* ( NeurIPS'21 )~\cite{zhang2021flexmatch} & 61.78$\pm$ 1.17 &  + \textbf{Our PCL} &  \textbf{64.15$\pm$0.53} \\     
					\hline
					\specialrule{.1em}{.05em}{.05em}
				\end{tabular}}
    \vspace{0.3 mm}
    \caption{Classification accuracy (\%) of SSL for CIFAR-100 (400 labels). * means our reimplementation.} 
\label{table:ssl_cifar}
\end{table}

In fact, as long as the features of unlabeled data cannot be clustered around the class weights, PCL has good potential to improve the performance.
In this section, we consider the case where the source domain and target domain come from the same distribution, \textit{i.e.}, semi-supervised learning.
In particular, for the semi-supervised tasks, the unlabeled features will deviate from the class weight when the labeled data is very scarce.
Therefore, we consider the case where there are only 4 labeled samples per class. 

\noindent\textbf{\textit{Setup}} We conduct the experiments on CIFAR-100 \cite{krizhevsky2009learning} and take three SSL methods, including Fixmatch~\cite{fixmatch}, CCSSL~\cite{yang2022class} and Flexmatch~\cite{zhang2021flexmatch} as our baseline. For the hyperparameter in PCL, we set $s=7$ in all experiments.  

The evaluation results are reported in Table~\ref{table:ssl_cifar}.
For these three different baselines, PCL can bring significant gains.
This further proves our conclusion that PCL has the potential to improve model performance in scenarios where feature and class weights deviate.



\section{Discussion and Conclusion}

In this paper, we propose a simple yet effective probabilistic contrastive learning to address the problem of feature deviation from weights. Therefore, our method has shown positive results on multiple domain adaptation tasks. An open question worth discussing is: Can PCL be applied into general contrastive learning (GCL) for classification? Due to the absence of a classifier in the unsupervised pre-training stage of GCL, directly applying PCL meets challenges. A feasible solution is to employ a clustering algorithm to construct class centers and then use PCL to learn more robust features. We hope that PCL can bring some useful insights into general unsupervised representation learning tasks.


\section*{Acknowledgments}

This work is jointly supported by the National Natural
Science Foundation of China under Grant (62176246, 62276031) and Anhui Provincial Key Research and Development Project (202304a05020045).

{
\bibliographystyle{named}
\bibliography{ijcai24}
}
\def\x{{\mathbf x}}
\def\L{{\cal L}}


		\begin{table*}[h]			
			\begin{center}
      \resizebox{1.0\textwidth}{!}{
				\begin{tabular}{ll|ccccccccccccccccccc|c}
					
					& Method  & \rotatebox{90}{road} & \rotatebox{90}{sdwk} & \rotatebox{90}{bld} & \rotatebox{90}{wall} & \rotatebox{90}{fnc} & \rotatebox{90}{pole} & \rotatebox{90}{lght} & \rotatebox{90}{sign} & \rotatebox{90}{veg.} & \rotatebox{90}{trrn.} & \rotatebox{90}{sky} & \rotatebox{90}{pers} & \rotatebox{90}{rdr} & \rotatebox{90}{car} & \rotatebox{90}{trck} & \rotatebox{90}{bus} & \rotatebox{90}{trn} & \rotatebox{90}{mtr} & \rotatebox{90}{bike} & mIoU \\ \toprule    
					\midrule 

					&  DACS (WACV'20)~\cite{tranheden2021dacs}  & 89.9  & 39.7 & 87.9 & 30.7 & 39.5 & 38.5 & 46.4 & 52.8 & 88.0 & 44.0 & 88.8 & 67.2 & 35.8 & 84.5 & 45.7 & 50.19 & 0.0 & 27.3 & 34.0 & 52.1 \\
					
					& SPCL (arXiv'21) ~\cite{xie2021spcl} & 90.3  & 50.3 & 85.7 & 45.3 & 28.4 & 36.8 & 42.2 & 22.3 & 85.1 & 43.6 & 87.2 & 62.8 & 39.0 & 87.8 & 41.3 & 53.9 & 17.7 & 35.9 & 33.8 & 52.1 \\
					
					& RCCR (arXiv'21)~\cite{zhou2021domain}   & 93.7  & 60.4 & 86.5 & 41.1 & 32.0 & 37.3 & 38.7 & 38.6 & 87.2 & 43.0 & 85.5 & 65.4 & 35.1 & 88.3 & 41.8 & 51.6 & 0.0 & 38.0 & 52.1 & 53.5 \\
					
					&  SAC (arXiv'21)~\cite{araslanov2021self}   & 90.4 & 53.9 & 86.6 & 42.4 & 27.3 & 45.1 & 48.5 & 42.7 & 87.4 & 40.1 & 86.1 & 67.5 & 29.7 & 88.5 & 49.1 & 54.6 & 9.8 & 26.6 & 45.3 & 53.8 \\ 
					& Pixmatch (CVPR'21)~\cite{melas2021pixmatch}   & 91.6  & 51.2 & 84.7 & 37.3 & 29.1 & 24.6 & 31.3 & 37.2 & 86.5 & 44.3 & 85.3 & 62.8 & 22.6 & 87.6 & 38.9 & 52.3 & 0.7 & 37.2 & 50.0 & 50.3 \\
     &  ProDA (CVPR'2021)~\cite{zhang2021prototypical}   & - & - & - & - & - & - & - & - & - & - & - & - & - & - & - & - & - & - & - & 53.7 \\
					& DPL (ICCV'21)~\cite{cheng2021dual}   & 92.8  & 54.4 & 86.2 & 41.6 & 32.7 & 36.4 & 49.0 & 34.0 & 85.8 & 41.3 & 83.0 & 63.2 & 34.2 & 87.2 & 39.3 & 44.5 & 18.7 & 42.6 & 43.1 & 53.3 \\
					& BAPA (ICCV'21)~\cite{liu2021bapa}  & 94.4  & 61.0 & 88.0 & 26.8 & 39.9 & 38.3 & 46.1 & 55.3 & 87.8 & 46.1 & 89.4 & 68.8 & 40.0 & 90.2 & 60.4 & 59.0 & 0.00 & 45.1 & 54.2 & 57.4 \\
					& UPST (ICCV'21)~\cite{wang2021uncertainty}  & 90.5  & 38.7 & 86.5 & 41.1 & 32.9 & 40.5 & 48.2 & 42.1 & 86.5 & 36.8 & 84.2 & 64.5 & 38.1 & 87.2 & 34.8 & 50.4 & 0.2 & 41.8 & 54.6 & 52.6 \\
					& DSP (ACM MM'21)~\cite{gao2021dsp}   & 92.4  & 48.0 & 87.4 & 33.4 & 35.1 & 36.4 & 41.6 & 46.0 & 87.7 & 43.2 & 89.8 & 66.6 & 32.1 & 89.9 & 57.0 & 56.1 & 0.0 & 44.1 & 57.8 & 55.0 \\ 
				& CPSL (CVPR'22)~\cite{li2022class}   & 91.7 & 52.9 & 83.6 & 43.0 & 32.3 & 43.7 & 51.3 & 42.8 & 85.4 & 37.6 & 81.1 & 69.5 & 30.0 & 88.1 & 44.1 & 59.9 & 24.9 & 47.2 & 48.4 &  55.7 \\

					\midrule
					\midrule 
					&  MFA-D (BMVC'21)~\cite{zhang2021multiple}   & 93.5 & 61.6 & 87.0 & 49.1 & 41.3 & 46.1 & 53.5 & 53.9 & 88.2 & 42.1 & 85.8 & 71.5 & 37.9 & 88.8 & 40.1 & 54.7 & 0.0 & 48.2 & 62.8 & 58.2 \\
					
					&  ProDA-D (CVPR'21)~\cite{zhang2021prototypical}   & 87.8 & 56.0 & 79.7 & 46.3 & 44.8 & 45.6 & 53.5 & 53.5 & 88.6 & 45.2 & 82.1 & 70.7 & 39.2 & 88.8 & 45.5 & 59.4 & 1.0 & 48.9 & 56.4 & 57.5 \\
					
					& ProDA-D + CaCo (CVPR'22)~\cite{huang2022category}  & 93.8 & 64.1 & 85.7 & 43.7 & 42.2 & 46.1 & 50.1 & 54.0 & 88.7 & 47.0 & 86.5 & 68.1 & 2.9 & 88.0 & 43.4 & 60.1 & 31.5 & 46.1 & 60.9 & 58.0 \\					
					& ProDA-D + CRA (arXiv'21)~\cite{wang2021cross}   & 89.4 & 60.0 & 81.0 & 49.2 & 44.8 & 45.5 & 53.6 & 55.0 & 89.4 & 51.9 & 85.6 & 72.3 & 40.8 & 88.5 & 44.3 & 53.4 & 0.0 & 51.7 & 57.9 &  58.6 \\
					& CPSL-D (CVPR'2022)~\cite{li2022class}   & 92.3 & 59.9 & 84.9 & 45.7 & 29.7 & 52.8 & 61.5 & 59.5 & 87.9 & 41.5 & 85.0 & 73.0 & 35.5 & 90.4 & 48.7 & 73.9 & 26.3 & 53.8 & 53.9 &  60.8 \\     						
					\midrule \midrule 
     					& BAPA*  & 94.1 & 61.5 & 87.6 & 38.4 & 37.9 & 37.4 & 46.7 & 59.2 & 88.0 & 47.0 & 88.3 & 69.0 & 32.1 & 91.2 & 62.8 & 61.7 & 0.6 & 42.9 & 49.9 &  57.7 \\
					& +  \textbf{Our PCL}  & 95.6 & 71.0  & 89.1 & 46.8 & 43.4 & 45.8 & 53.5 & 64.1 & 88.3 & 43.3 & 87.6 & 70.9 & 44.5 & 89.8 & 57.9 & 61.5 & 0.0 & 42.4 & 58.5 & \bf 60.7 \\
										
					\bottomrule
				\end{tabular}}

			\end{center}
   				\caption{Results on the GTA5$\rightarrow$Cityscapes benchmark. D means using distillation technique.}
       \label{tab:gta}
		\end{table*}

	\begin{table*}[h]

		\begin{center}
			\addtolength{\leftskip} {-1.3cm} 
			\addtolength{\rightskip}{-1.8cm}
			
			\vspace{1mm}
           \resizebox{1.0\textwidth}{!}{
			\begin{tabular}{ll|cccccccccccccccc|c|c}
				& Method   & \rotatebox{90}{road} & \rotatebox{90}{sdwk} & \rotatebox{90}{bld} & \rotatebox{90}{wall$^{*}$} & \rotatebox{90}{fnc$^{*}$} & \rotatebox{90}{pole$^{*}$} & \rotatebox{90}{light} & \rotatebox{90}{sign} & \rotatebox{90}{veg.} & \rotatebox{90}{sky} & \rotatebox{90}{pers} & \rotatebox{90}{rdr} & \rotatebox{90}{car} & \rotatebox{90}{bus} & \rotatebox{90}{mtr} & \rotatebox{90}{bike} & mIoU-16 & mIoU-13 \\ \toprule
				\midrule 
				&  DACS (WACV'20)~\cite{tranheden2021dacs}  & 80.6 & 25.1 & 81.9 & 21.5 & 2.9 & 37.2 & 22.7 & 24.0 & 83.7 & 90.8 & 67.6 & 38.3 & 82.9 & 38.9 & 28.5 & 47.6 & 48.3 & 54.8 \\ 
				&  SPCL (arXiv'21)~\cite{xie2021spcl} & 86.9 & 43.2 & 81.6 & 16.2 & 0.2 & 31.4 & 12.7 & 12.1 & 83.1 & 78.8 & 63.2 & 23.7 & 86.9 & 56.1 & 33.8 & 45.7 & 47.2 & 54.4 \\
				&  RCCR (arXiv'21)~\cite{zhou2021domain}   &  79.4 & 45.3 & 83.3 & - & - & - & 24.7 & 29.6 & 68.9 & 87.5 & 63.1 & 33.8 & 87.0 & 51.0 & 32.1 & 52.1 & - & 56.8 \\			
				&  SAC (CVPR'21)~\cite{araslanov2021self}    &  89.3 & 47.2 & 85.5 & 26.5 & 1.3 & 43.0 & 45.5 & 32.0 & 87.1 & 89.3 & 63.6 & 25.4 & 86.9 & 35.6 & 30.4 & 53.0 & 52.6 & 59.3 \\
				&  Pixmatch (CVPR'21)~\cite{melas2021pixmatch}  & 92.5 & 54.6 & 79.8 & 4.8 & 0.1 & 24.1 & 22.8 & 17.8 & 79.4 & 76.5 & 60.8 & 24.7 & 85.7 & 33.5 & 26.4 & 54.4 & 46.1 & 54.5 \\
				& DPL (ICCV'21)~\cite{cheng2021dual}  & 87.5 & 45.7 & 82.8 & 13.3 & 0.6 & 33.2 & 22.0 & 20.1 & 83.1 & 86.0 & 56.6 & 21.9 & 83.1 & 40.3 & 29.8 & 45.7 & 47.0 & 54.2 \\
				& BAPA (ICCV'21)~\cite{liu2021bapa}  & 91.7 & 53.8 & 83.9 & 22.4 & 0.8 & 34.9 & 30.5 & 42.8 & 86.6 & 88.2 & 66.0 & 34.1 & 86.6 & 51.3 & 29.4 & 50.5 & 53.3 & 61.2 \\
				& UPST (ICCV'21)~\cite{wang2021uncertainty}  & 79.4 & 34.6 & 83.5 & 19.3 & 2.8 & 35.3 & 32.1 & 26.9 & 78.8 & 79.6 & 66.6 & 30.3 & 86.1 & 36.6 & 19.5 & 56.9 & 48.0 & 54.6 \\					
				& DSP (ACM MM'21)~\cite{gao2021dsp}   & 86.4 & 42.0 & 82.0 & 2.1 & 1.8 & 34.0 & 31.6 & 33.2 & 87.2 & 88.5 & 64.1 & 31.9 & 83.8 & 65.4 & 28.8 & 54.0 & 51.0 & 59.9 \\
& CPSL-D (CVPR'22)~\cite{li2022class}  & 87.2 & 43.9 & 85.5 & 33.6 & 0.3 & 47.7 & 57.4 & 37.2 & 87.8 & 88.5 & 79.0 & 32.0 & 90.6 & 49.4 & 50.8 & 59.8 & 57.9 & 65.3  \\

				\midrule 
				\midrule 
				&  MFA-D  (BMVC'2021)  ~\cite{zhang2021multiple} & 81.8 & 40.2 & 85.3 & - & - & - & 38.0 & 33.9 & 82.3 & 82.0 & 73.7 & 41.1 & 87.8 & 56.6 & 46.3 & 63.8 & - & 62.5 \\
				
				& ProDA-D  (CVPR'2021)~\cite{zhang2021prototypical}  & 87.8 & 45.7 & 84.6 & 37.1 & 0.6 & 44.0 & 54.6 & 37.0 & 88.1 & 84.4 & 74.2 & 24.3 & 88.2 & 51.1 & 40.5 & 45.6 & 55.5 & 62.0 \\
				
				& ProDA-D + CRA  (arXiv'2021)~\cite{wang2021cross}  & 85.6 & 44.2 & 82.7 & 38.6 & 0.4 & 43.5 & 55.9 & 42.8 & 87.4 & 85.8 & 75.8 & 27.4 & 89.1 & 54.8 & 46.6 & 49.8 & 56.9 & 63.7 \\
					& CPSL-D (CVPR'2022) ~\cite{li2022class}  & 87.2 & 43.9 & 85.5 & 33.6 & 0.3 & 47.7 & 57.4 & 37.2 & 87.8 & 88.5 & 79.0 & 32.0 & 90.6 & 49.4 & 50.8 & 59.8 & 57.9 & 65.3  \\				
				\midrule \midrule 
				& BAPA*  & 71.0 & 36.8 & 76.8 & 26.0 & 4.2 & 40.8 & 39.1 & 35.4 & 88.7 & 88.6 & 69.7 & 34.3 & 89.1 & 62.8 & 43.7 & 45.3 &  53.3 &  60.1 \\    
				& + \textbf{Our PCL}  & 91.2 & 63.1 & 84.7 & 30.3 & 6.7 & 42.2 & 47.4 & 49.6 & 89.4 & 91.3 & 70.2 & 39.2 & 91.4 & 66.8 & 44.7 & 57.1 & \bf 60.3 & \bf 68.2 \\
				
				\bottomrule
			\end{tabular}}
			
		\end{center}  
  			\caption{Results on the SYNTHIA$\rightarrow$Cityscapes benchmark. mIoU-16 and mIoU-13 refer to mean intersection-over-union on the standard sets of 16 and 13 classes, respectively. Classes not evaluated are replaced by '*'. D means using distillation technique.}
     \label{tab:synthia}
	\end{table*}
\newpage
\section{Supplementary Material}
In the supplementary material, we present the implementation details of each domain adaptation task and more experimental results.

\subsection{Implementation Details}

\subsubsection{UDA Semantic Segmentation}
Following the widely used  UDA Semantic segmentation protocols~\cite{tranheden2021dacs,melas2021pixmatch,zhou2021domain,liu2021bapa}, we use the DeepLab-v2 segmentation model~\cite{chen2017deeplab} with a ResNet-101~\cite{ResNet1} backbone pre-trained on ImageNet as our model. Following~\cite{melas2021pixmatch}, we first perform warm-up training on source domain. We adopt SGD with momentum of 0.9 and set the weight decay to $5\times10^{-4}$. The learning rate is set at $2.5\times10^{-4}$ for backbone and $2.5\times10^{-3}$ for others. During training we use a poly policy with an exponent of 0.9 for learning rate decay. The max iteration number is set to 250k. For the hyper-parameters in PCL, we set $s=20$ in all experiments. The data augmentation used by PCL is consistent with BAPA.

\textbf{Cityscapes}~\cite{cordts2016cityscapes} is a representative dataset in semantic segmentation and autonomous driving domain, and we use it as the target domain dataset. The training and validation subsets contain $2,975$ and $500$ annotated images with $2048\times1024$ resolution taken from real urban street scenes.
\textbf{SYNTHIA}~\cite{ros2016synthia} is also a synthetic dataset consisting of $9,400$ annotated images with $1280\times720$ resolution, which is used as another source domain dataset.
 
\subsubsection{UDA Detection}
\label{app:DAOD}
Following~\cite{zhang2021rpn}, we adopt the Faster R-CNN~\cite{ren2015faster} with VGG16~\cite{simonyan2014very} as backbone and initialized by the model pre-trained on ImageNet. We follow the original papers~\cite{zhang2021rpn} and train the model for 80k iterations. The training strategies of detector and discriminator are also consistent with RPN. For detector, we use SGD to train it. For the first 50k iterations, we set the learning rate to 0.001, and then we drop it to 0.0001 for the last 30k.
For discriminators, we adopt Adam optimizer~\cite{kingma2014adam} with a learning rate of 0.0001. For the hyper-parameters in PCL, we set $s=20$. The data augmentation used by PCL is consistent with RPN and FixMatch.

\textbf{\textit{Sim10k}}~\cite{SIM10K} is a collection of synthesized images that contains 10, 000 images and corresponding bounding box annotations. In cross-domain detection tasks, Sim10K is usually used as the source domain and Cityscapes is used as the target domain for experiments.

\subsubsection{UDA}
\label{app:uda}
Following the standard transductive setting~\cite{cui2020gradually} for UDA, we use all labeled source data and all unlabeled target data, and test on the same unlabeled target data. For model selection, we use ResNet-50~\cite{ResNet1} pre-trained on ImageNet as the backbone network for both Office-Home and VisDA-2017. For training hyper-parameters, we use mini-batch stochastic gradient descent (SGD) with a momentum of 0.9 and a weight decay of 0.001. For Office-Home, the initial learning rate is set to 0.001. For VisDA-2017, an initial learning rate of 0.0003 is used. The max iteration number is set to 10k.

When applying FixMatch in domain adaptation task, there exist wrong predicted high confident target samples, which can hurt the performance in the target domain. To amend this, we follow~\cite{zhang2021prototypical,li2021ecacl} which uses a regularization term from~\cite{zou2019crst}. It encourages the high confident output to be evenly distributed to all classes. For the hyper-parameters in PCL, we set $s=7$ in all experiments. The data augmentation used by PCL is consistent with GVB and FixMatch.

\begin{equation}
    \ell_{reg} = -\sum_{i=1}^{N}\sum_{j=1}^{C} \frac{1}{C} \log p_t^{(i,j)}.
\label{eq:kld}
\end{equation}
where $N$ denotes for the number of high confident samples, $C$ denotes for the number of classes.

 \textit{\textbf{Office-Home}}~\cite{Office-HOME} consists of images of everyday objects organized into four domains: Artistic (Ar), Clipart (Cl), Product (Pr), and Real-world (Rw). It contains 15,500 images of 65 classes.  \textit{\textbf{VisDA-2017}}~\cite{VisDA2017} is a large-scale dataset for synthetic-to-real domain adaptation. It contains 152,397 synthetic images for the source domain and 55,388 real-world images for the target domain. 
 
\subsubsection{SSDA}
\label{SSDA-detail}

Following MME~\cite{saito2019semi}, we remove the last linear layer of AlexNet and ResNet34, while adding a new classifier $F$. We also use the model pre-trained on ImageNet to initialize all layers except $F$. We adopt SGD with momentum of 0.9 and set the initial learning rate is 0.01 for fully-connected layers whereas it is set 0.001 for other layers. The max iteration number is set to 50k. In SSDA task, we also use the regularization term in Equation~(\ref{eq:kld}). For the hyper-parameters in PCL, we set $s=7$ in all experiments.
 The data augmentation used by PCL is consistent with MME and FixMatch.

\textbf{\textit{DomainNet}} is initially a multi-source domain adaptation benchmark. Similar to MME~\cite{saito2019semi}, we only select 4 domains Real, Clipart, Painting, and Sketch~(abbr. {\bf R}, {\bf C}, {\bf P}, and {\bf S}), each of which contains images of 126 categories. \textbf{\textit{Office-Home}} is a widely used UDA benchmark and consists of Real, Clipart, Art, and Product~(abbr. {\bf R}, {\bf C}, {\bf A}, and {\bf P}) domains with 65 classes.

\subsubsection{SSL}
\label{app:ssl}
Following~\cite{fixmatch,zhang2021flexmatch,yang2022class}, we report the performance of an EMA model and use a WRN-28-8 for Cifar100. We follow the original papers~\cite{fixmatch,zhang2021flexmatch,yang2022class}, the model is trained using SGD with a momentum of 0.9  and using an learning rate of 0.03 with a cosine decay schedule.
 For Fixmatch~\cite{fixmatch} and Flexmatch~\cite{zhang2021flexmatch}, the model is trained for 1024 epochs. For CCSSL~\cite{yang2022class}, the model is trained for 512 epochs. For the hyper-parameters in PCL, we set $s=7$ in all experiments. The data augmentation used by PCL is completely consistent with each baseline.

 \textit{\textbf{CIFAR-100}}~\cite{krizhevsky2009learning} contains 50,000 images of size $32\times32$ from 100 classes. To ensure fairness, we strictly follow the standard data partitioning strategy~\cite{zhang2021flexmatch}.
\subsection{More Results}
\subsubsection{More Results on UDA Semantic Segmentation}
Table~\ref{tab:gta} and Table~\ref{tab:synthia}  give the detailed results of different methods for each category on GTA5~\cite{richter2016playing}$\rightarrow$Cityscapes~\cite{cordts2016cityscapes} and SYNTHIA~\cite{ros2016synthia}$\rightarrow$Cityscapes.
It can be seen that our method can significantly improve the performance of the baseline.

\subsubsection{More Results on UDA}
\begin{table*}[ht]
  	
		\resizebox{\textwidth}{!}{
			\begin{tabular}{l@{}|cccccccccccc|c}
				\hline
				Method& A$\rightarrow$C & A$\rightarrow$P & A$\rightarrow$R & C$\rightarrow$A & C$\rightarrow$P & C$\rightarrow$R & P$\rightarrow$A & P$\rightarrow$C & P$\rightarrow$R & R$\rightarrow$A & R$\rightarrow$C & R$\rightarrow$P & Avg \\
				\hline
				Source-Only  & 34.9 & 50.0 & 58.0 & 37.4 & 41.9 & 46.2 & 38.5 & 31.2 & 60.4 & 53.9 & 41.2 & 59.9 & 46.1 \\
				TAT (ICML'19)~\cite{liu2019transferabletat} & 51.6 & 69.5 & 75.4 & 59.4 & 69.5 & 68.6 & 59.5 & 50.5 & 76.8 & 70.9 & 56.6 & 81.6 & 65.8 \\
				SymNet (CVPR'19)~\cite{zhang2019domainsymnet} & 47.7 & 72.9 & 78.5 & 64.2 & 71.3 & 74.2 & 63.6  & 47.6 & 79.4 & 73.8 & 50.8 & {82.6} & 67.2 \\
				MDD (ICML'19)~\cite{zhang2019bridgemdd} &
				54.9 & 73.7 & 77.8 & 60.0 & 71.4 &71.8 & 61.2 & 53.6 & 78.1 & 72.5 & 60.2 & 82.3 & 68.1 \\
				BNM (CVPR'20)~\cite{cui2020towardsbnm}
				& 56.2 & 73.7 & 79.0 & 63.1 & 73.6 & 74.0 & 62.4 & 54.8 & 80.7 & 72.4 & 58.9 & 83.5 & 69.4 \\
				FixBi (CVPR'21)~\cite{na2021fixbi} & 58.1 & 77.3 & 80.4 & 67.7 & \textbf{79.5} & 78.1 & 65.8 & 57.9 & 81.7 & 76.4 & 62.9 & \textbf{86.7} & 72.7  \\
    RADA  (CVPR'21)~\cite{jin2021re} & 56.5 & 76.5 & 79.5 & 68.8 & 76.9 & 78.1 & 66.7 & 54.1 & 81.0 & 75.1 & 58.2 & 85.1 & 71.4  \\
				ToAlign (NeurIPS'21)~\cite{wei2021toalign} & 57.9 & 76.9 & 80.8 & 66.7 & 75.6 & 77.0 & 67.8 & 57.0 & 82.5 & 75.1 & 60.0 & 84.9 & 72.0  \\              CST (NeurIPS'21)~\cite{liu2021cycle} & 59.0 & 79.6 & 83.4 & 68.4 & 77.1 & 76.7 & 68.9 & 56.4 & 83.0 & 75.3 & 62.2 & 85.1 & 73.0  \\   
				SCDA (ICCV'21)~\cite{li2021semantic} & 60.7 & 76.4 & 82.5 & 69.8 & 77.5 & 78.4 & 68.9 & 59.0 & 82.7 & 74.9 & 61.8 & 84.5 & 73.1  \\				
				TCM (ICCV'21)~\cite{yue2021transporting} & 58.6 & 74.4 & 79.6 & 64.5 & 74.0 & 75.1 & 64.6 & 56.2 & 80.9 & 74.6 & 60.7 & 84.7 & 70.7  \\				 SDAT (ICML'22)~\cite{rangwani2022closer} & 58.2 & 77.1 & 82.2& 66.3 & 77.6 & 76.8 & 63.3 & 57.0 & 82.2 & 74.9 & 64.7 & 86.0 & 72.2 \\	
			 SDAT (ICML'22)~\cite{rangwani2022closer} & 58.2 & 77.1 & 82.2& 66.3 & 77.6 & 76.8 & 63.3 & 57.0 & 82.2 & 74.9 & 64.7 & 86.0 & 72.2 \\	   
    				NWD (CVPR'22)~\cite{chen2022reusing} & 58.1 & 79.6 & \textbf{83.7} & 67.7 & 77.9 & 78.7 & 66.8 & 56.0 & 81.9 & 73.9 & 60.9 & 86.1 & 72.6  \\	
        HMA ( ICCV'23 )~\cite{zhou2023homeomorphism} & 60.6 & 79.1 & 82.9 & 68.9 & 77.5 & \textbf{79.3} & 69.1 & 55.9 & 83.5 & 74.6 & 62.3 & 84.4 & 73.2  \\	
				\hline
				\hline
				GVB* (CVPR'20)~\cite{cui2020gradually} & 57.4 & 74.9 & 80.1 & 64.1 & 73.9 & 74.3 & 65.0 &  55.9 & 81.1 & 75.0 & 58.1 & 84.0 & 70.3 \\
				+ MetaAlign(CVPR'21)~\cite{wei2021metaalign} & $59.3$ & $76.0$ & $80.2$ & $65.7$ & $74.7$ & $75.1$ & $65.7$ & $56.5$ & $81.6$ & $74.1$ & $ 61.1 $ & $85.2$ & $71.3$ \\
				+ \textbf{Our PCL} & $59.7$ & $75.9$ & $80.4$ & $69.3$ & $75.5$ & $77.1$ & $67.0$ & $58.3$ & $81.0$ & $75.2$ & $ 63.9 $ & $84.6$ & $72.3$ \\
				\hline
				\hline                
				GVB$^\dag$ & 59.8 & 78.1 & 81.3 & 67.7 & 78.2 & 76.7 & 68.7 &  60.2 & \textbf{83.9} & 75.1 & 65.5 & 86.4 & 73.5 \\
				+ \textbf{Our PCL} & \textbf{60.8} & \textbf{79.8} & $81.6$ & $\textbf{70.1}$ & $78.9 $ & $78.9$ & $\textbf{69.9}$ & $\textbf{60.7}$ & $83.3$ & \textbf{77.1} & $ \textbf{66.4} $ & $85.9$ & $\textbf{74.5}$ \\				
				\hline
		\end{tabular}}
 
	\caption{
	Accuracy (\%) of different UDAs on Office-Home with ResNet-50.}
\label{tab:uda_officehome_app}  
\end{table*}
The \textbf{Office-Home}~\cite{Office-HOME} dataset in the UDA scenario contains 12 UDA tasks, and we give detailed performance for each UDA task in Table~\ref{tab:uda_officehome_app}. It can be seen that PCL has obvious gains on both GVB* and GVB$^\dag$.

\subsubsection{More Results on SSDA}
\begin{table*}[htbp]

\fontsize{7.5pt}{7.5pt} 
\selectfont
\begingroup
\setlength{\tabcolsep}{1.2pt} 
\renewcommand{\arraystretch}{1.1} 
\begin{center}
\resizebox{\linewidth}{!}{
\begin{tabular}{c|l@{}|cccccccccccccc|cc}
\specialrule{.1em}{.05em}{.05em}
\multirow{2}{*}{Net} & \multirow{2}{*}{Method} & \multicolumn{2}{c}{R$\rightarrow$C} & \multicolumn{2}{c}{R$\rightarrow$P} & \multicolumn{2}{c}{P$\rightarrow$C} & \multicolumn{2}{c}{C$\rightarrow$S} & \multicolumn{2}{c}{S$\rightarrow$P} & \multicolumn{2}{c}{R$\rightarrow$S} & \multicolumn{2}{c|}{P$\rightarrow$R} & \multicolumn{2}{c}{Mean} \\
 & & 1-shot & 3-shot & 1-shot & 3-shot & 1-shot & 3-shot & 1-shot & 3-shot & 1-shot & 3-shot & 1-shot & 3-shot & 1-shot & 3-shot & 1-shot & 3-shot \\ \hline
\hline
\multirow{10}{*}{R} 
 & CLDA (NeurIPS'21)~\cite{singh2021clda} & 76.1 & 77.7 & 75.1 & 75.7 & 71.0 & 76.4 & 63.7& 69.7 & 70.2 & 73.7 & 67.1 & 71.1 & 80.1 & 82.9 & 71.9 & 75.3 \\
  & MME$^{*}$~\cite{saito2019semi} & 71.0 & 71.4 & 68.9 & 70.0 & 69.2 & 72.6 & 59.8 &62.7 & 65.6 & 68.2 & 63.2 & 64.3 & 77.8 & 77.9 & 67.9 & 69.5   \\
  & + \textbf{Our PCL} & 74.8 & 78.1 & 73.9 & 76.5 & 75.5 & 78.6 & 67.6 & 72.5 & 73.4 & 75.6 & 68.9 & 72.5 & 80.6 & \textbf{84.6} & 73.5 & 76.9   \\  
\cline{2-18}  

 &MCL$^\dag$ ( IJCAI'22 )~\cite{yan2022multi} &  77.4 &  79.4 &  74.6 &  76.3 & 75.5 & 78.8 & 66.4 & 70.9 &  74.0 &  74.7 &  70.7 &  72.3 &  82.0 &  83.3 &  74.4 &  76.5  \\
 &ProMM$^\dag$ ( IJCAI'23 )~\cite{huang2023semi} &  78.5 &  80.2 &  75.4 &  76.5 & \textbf{77.8} & 78.9 & \textbf{70.2} & 72.0 &  74.1 &  75.4 &  \textbf{72.4} &  \textbf{73.5} &  \textbf{84.0} &  \textbf{84.8} &  \textbf{76.1} &  77.4  \\
  & ECACL-P$^\dag$ (ICCV'21)~\cite{li2021ecacl} &  75.3 & 79.0 & 74.1 &  77.3 &  75.3 &  79.4 &  65.0 &  70.6 &  72.1 &  74.6 &  68.1 &  71.6 &  79.7 &  82.4 &  72.8 & 76.4 \\  
  \cdashline{1-4}[0.8pt/2pt]
 & CDAC$^\dag$ (CVPR'21)~\cite{li2021cdac} &  77.4 &  79.6 &  74.2 &  75.1 & 75.5 & 79.3 & 67.6 & 69.9 &  71.0 &  73.4 &  69.2 &  72.5 &  80.4 &  81.9 &  73.6 &  76.0 \\
     &+SLA ( CVPR'23 )~\cite{yu2023semi} &  \textbf{79.8} & \textbf{81.6} &  \textbf{75.6} &  76.0 & 77.4 & \textbf{80.3} & 68.1 & 71.3 &  71.7 &  73.5 &  71.7 &  \textbf{73.5} &  80.4 &  82.5 &  75.0 &  76.9 \\
   \cdashline{1-4}[0.8pt/2pt]    
  &MME$^\dag$ &  75.5 &  78.7 &  72.5 &  77.0 &  75.9 &  80.0 &  66.3 &  68.6 & 72.1 &  74.4 &  67.2 &  71.4 &  81.1 &  82.6 & 72.9 &  76.1   \\
 & + \textbf{Our PCL}  & 78.1 & 80.5 & 75.2 & \textbf{78.1} & 77.2 & \textbf{80.3} & 68.8 & \textbf{74.1} & \textbf{74.5} & \textbf{76.5} & 70.1 & \textbf{73.5} & 81.9 & 84.1 & 75.1 & \textbf{78.2}  \\
 \specialrule{.1em}{.05em}{.05em}

\end{tabular}}
\label{tab:ssda_domainnet_all}
\end{center}
\endgroup
\caption{Accuracy(\%) on \textit{DomainNet} under the settings of 1-shot and 3-shot using Alexnet (A) and Resnet34 (R) as backbone networks. ${\dag}$ means using Fixmath and ${*}$ means our reimplementation.}
\end{table*}
\begin{table*}[ht]
\centering
\resizebox{\linewidth}{!}{
\begin{tabular}{c|l@{}|cccccccccccc|c}
\specialrule{.1em}{.05em}{.05em}
Net & Method & R$\rightarrow$C & R$\rightarrow$P & R$\rightarrow$A & P$\rightarrow$R & P$\rightarrow$C & P$\rightarrow$A & A$\rightarrow$P & A$\rightarrow$C & A$\rightarrow$R & C$\rightarrow$R & C$\rightarrow$A & C$\rightarrow$P & Mean \\
\hline
\cline{2-15} 
\hline
\hline
\multirow{7}{*}{R} 

  & CLDA (NeurIPS'21)~\cite{singh2021clda} & 66.0 & 87.6 & 76.7 & 82.2 & 63.9 & 72.4 & 81.4 & 63.4 & 81.3 & 80.3 & 70.5 & 80.9 & 75.5 \\

  & MME$^{*}$~\cite{saito2019semi} & 66.0 & 86.0 & 72.3 & 80.4 & 64.0 & 67.4 & 79.8 & 64.0 & 77.9 & 77.1 & 66.6 & 80.0 & 73.5 \\
  & + \textbf{PCL} & 65.4 & 86.7 & 74.5 & 83.1 & 62.9 & 71.0 & 82.8 & 63.7 & 81.0 & 81.1 & 71.0 & 83.1 & 75.5 \\
  \cline{2-15}
&MCL$^\dag$ ( IJCAI'22 )~\cite{yan2022multi} & 70.1 & 88.1 & 75.3 & 83.0 & 68.0 & 69.9 & 83.9 & 67.5 & 82.4 & 81.6 & 71.4 & 84.3 & 77.1 \\
    &ProMM$^\dag$ ( IJCAI'23 )~\cite{huang2023semi} & \textbf{71.0} & 88.6 & 75.8 & \textbf{83.8} & \textbf{68.9} & 72.5 & 83.9 & \textbf{67.8} & 82.2 & 82.3 & 72.1 & \textbf{84.1} & 77.8 \\
  \cdashline{1-4}[0.8pt/2pt]
 & CDAC$^\dag$ (CVPR'21)~\cite{li2021cdac} & 67.8 & 85.6 & 72.2 & 81.9 & 67.0 & 67.5 & 80.3 & 65.9 & 80.6 & 80.2 & 67.4 & 81.4 & 74.8 \\  
   &+SLA ( CVPR'23 )~\cite{yu2023semi} & 70.1 & 87.1 & 73.9 & 82.5 & 69.3 & 70.1 & 82.6 & 67.3 & 81.4 & 80.1 & 69.2 & 82.1 & 76.3  \\
       \cdashline{1-4}[0.8pt/2pt]
  & MME$^\dag$ & 67.2 & 88.1 &  76.6 & 83.0 & 66.8 & 73.6 & 83.8 & 67.3 & 80.5 & 81.1 & 71.8 & 82.7 & 76.9 \\

 & + \textbf{PCL}  & 69.1 & \textbf{89.5} & \textbf{76.9} & \textbf{83.8} & 68.0 & \textbf{74.7} & \textbf{85.5} & 67.6 & \textbf{82.3} & \textbf{82.7} & \textbf{73.4} & 83.4 & \textbf{78.1} \\ 
\specialrule{.1em}{.05em}{.05em}
\end{tabular}}
\caption{Accuracy(\%) on \textit{Office-Home} under the setting of 3-shot using Alexnet (A) and Resnet34 (R) as backbone networks. ${\dag}$ means using Fixmath and ${*}$ means our reimplementation.}
\label{tab:ssda_officehome_3_shot}
\vspace{-2.0mm}
\end{table*}
\textit{DomainNet}~\cite{peng2019moment} and \textit{Office-Home}.
Here we present the detailed results of each SSDA task on the \textit{DomainNet}~\cite{peng2019moment} and \textit{Office-Home}in Table~\ref{tab:ssda_domainnet_all} and Table~\ref{tab:ssda_officehome_3_shot}. In particular, we also conducted experiments with Alexnet~\cite{krizhevsky2012imagenet} as the backhone. It can be seen that PCL has obvious gains on both Alexnet and Resnet34.

\subsubsection{PCL v.s SPCL}
As discussed in Sec. 4.5, addressing false negtive samples problem can indeed improve the performance, however, the gain is quite limited. Similarly, if we further transform PCL into SPCL, the gain obtained is still very limited(76.9$\%$$\rightarrow$77.2$\%$). It further shows that in domain adaptation, mitigating the deviation between the features and class weights is more important than the false negative sample.

Since the gain brought by SPCL is limited and will make PCL more complex, considering the simplicity, we do not use SPCL as the final solution.

\subsubsection{PCL v.s. FCL on GVB}
\begin{table*}
	
		\resizebox{\textwidth}{!}{
			\begin{tabular}{l@{}|cccccccccccc|c}
				\hline
				Method& A$\rightarrow$C & A$\rightarrow$P & A$\rightarrow$R & C$\rightarrow$A & C$\rightarrow$P & C$\rightarrow$R & P$\rightarrow$A & P$\rightarrow$C & P$\rightarrow$R & R$\rightarrow$A & R$\rightarrow$C & R$\rightarrow$P & Avg \\
				\hline
				GVB* (CVPR'20)~\cite{cui2020gradually} & 57.4 & 74.9 & 80.1 & 64.1 & 73.9 & 74.3 & 65.0 &  55.9 & 81.1 & 75.0 & 58.1 & 84.0 & 70.3 \\
				+ FCL& $56.0$ & $75.3$ & $80.1$ & $65.7$ & $73.3$ & $75.2$ & $65.6$ & $54.4$ & $\textbf{81.5}$ & $74.4$ & $ 60.1 $ & $\textbf{84.7}$ & $70.5$ \\
				+ \textbf{Our PCL} & $\textbf{59.7}$ & $\textbf{75.9}$ & $\textbf{80.4}$ & $\textbf{69.3}$ & $\textbf{75.5}$ & $\textbf{77.1}$ & $\textbf{67.0}$ & $\textbf{58.3}$ & $81.0$ & $\textbf{75.2}$ & $ \textbf{63.9 }$ & $84.6$ & $\textbf{72.3}$\\
				\hline
		\end{tabular}}
	\caption{
	Classification accuracy (\%) of different methods on Office-Home with ResNet-50 as backbone.}    
\label{tab:uda_officehome_FLP}    
\end{table*}
In this part, we compare the effects of PCL, and FCL on GVB. Table~\ref{tab:uda_officehome_FLP} gives the results.
It can be seen that FCL can improve the performance of the baseline model, but it is very limited. Our PCL is obviously superior to FCL, which further proves the superiority of PCL.




\end{document}